\pdfoutput=1
\documentclass[11pt]{article}

\usepackage{ACL2023}

\usepackage{times}
\usepackage{latexsym}
\usepackage{amsmath}
\usepackage{amsfonts}
\usepackage{booktabs}
\usepackage{multirow}
\usepackage{cuted}
\usepackage{tikz}
\usepackage{pgfplots}
\pgfplotsset{compat=1.18}
\usepackage{float}
\usepackage{dblfloatfix}
\usepackage{colortbl}
\usepackage{siunitx}
\usepackage{enumitem}
\usepackage{listings}
\usepackage{xcolor}

\usepackage[T1]{fontenc}
\usepackage[utf8]{inputenc}

\usepackage{microtype}

\usepackage{inconsolata}
\usepackage{bm}

\title{Data Diversification Methods In Alignment Enhance Math Performance In LLMs}

\author{
\textbf{Berkan Dokmeci}$^{1}$,
\textbf{Qingyang Wu}$^{2}$,
\textbf{Ben Athiwaratkun}$^{2}$ \\
\textbf{Ce Zhang}$^{2,3}$,
\textbf{Shuaiwen Leon Song}$^{2}$,
\textbf{James Zou}$^{2,4}$ \\
$^1$Washington University in St. Louis,
$^2$Together AI,
$^3$University of Chicago,
$^4$Stanford University
}

\begin{document}
\maketitle
\begin{abstract}
While recent advances in preference learning have enhanced alignment in human feedback, mathematical reasoning remains a persistent challenge. We investigate how data diversification strategies in preference optimization can improve the mathematical reasoning abilities of large language models (LLMs). We evaluate three common data generation methods—temperature sampling, Chain-of-Thought prompting, Monte Carlo Tree Search (MCTS), and introduce Diversified-ThinkSolve (DTS), a novel structured approach that systematically decomposes problems into diverse reasoning paths. Our results show that with strategically diversified preference data, models can substantially improve mathematical reasoning performance, with the best approach yielding gains of 7.1\% on GSM8K and 4.2\% on MATH over the base model. Despite its strong performance, DTS incurs only a marginal computational overhead (1.03×) compared to the baseline, while MCTS is nearly five times more costly with lower returns. These findings demonstrate that structured exploration of diverse problem-solving methods creates more effective preference data for mathematical alignment than traditional approaches.
\end{abstract}

\section{Introduction}
Large language models (LLMs) have demonstrated remarkable capabilities across a wide range of tasks, but mathematical reasoning remains a particularly challenging domain \cite{luo2023wizardmath, lightman2023let}. While recent work has shown that Reinforcement Learning from Human Feedback (RLHF) \cite{stiennon2020learning} and preference optimization techniques like Direct Preference Optimization (DPO) \cite{rafailov2023direct} can substantially improve LLM performance on general tasks, their application to mathematical reasoning has received less attention. \nocite{xu2024dpo}

In standard preference optimization scenarios, datasets typically consist of unmodified preference pairs drawn from human annotations or model-generated evaluations. While such datasets can yield performance improvements \cite{guo2024direct-ipo, alignment_handbook2023, xia2024less}\nocite{chen2024alpagasus} in alignment with human preference, we hypothesize that more structured and diverse preference data can lead to significantly better performance specifically tailored to mathematical reasoning \cite{liu2024what}.

Our work explores how strategically designed data generation and diversification methods can enhance the effectiveness of preference optimization for mathematical reasoning. We propose several approaches to generate preference data that incorporate diverse reasoning strategies, problem reformulations, and solution methodologies. By leveraging techniques such as Chain-of-Thought (CoT) prompting \cite{wei2022chain, kojima2022large}, Monte Carlo Tree Search (MCTS) \cite{silver2016mastering, feng2023alphazero}, and specialized thought-reflection mechanisms, we create datasets that expose LLMs to a richer space of mathematical problem-solving strategies during preference optimization.

Among these approaches, we introduce Diversified-ThinkSolve (DTS), a novel structured method that systematically decomposes problems into diverse problem-solving approaches before generating solutions. DTS explicitly separates the thought generation process from solution execution, enabling exploration of multiple problem-solving strategies while maintaining computational efficiency. This approach addresses a fundamental limitation of traditional sampling methods—their inability to systematically explore diverse thinking pathways.

We conduct a comprehensive comparative analysis of these strategies across standard mathematics benchmarks. Our DTS approach yields significant improvements in both GSM8K and MATH over the base model, while incurring only marginal computational overhead. Our findings highlight that structured exploration of analytical approaches creates more effective preference data for mathematical alignment than traditional approaches, and that data quality and diversity can be more crucial than optimizing algorithmic approaches. 

% We propose several novel data diversification strategies for preference optimization tailored to mathematical reasoning tasks. 

% \qingyang{make sure the introduction is aligned with abstract and conclusion.}

\section{Background}
In this section, we provide the necessary background and information regarding alignment training for LLMs. We start by providing a background on the RLHF process and then we discuss post-training alignment techniques utilized in this paper. 

\subsection{Reinforcement Learning from Human Feedback}
Often after we pre-train a model we want to further adapt it to meet certain needs or specifications \cite{stiennon2020learning, bai2022training, Ouyang2022TrainingLM}\nocite{zhou2023lima}. Reinforcement Learning from Human Feedback (RLHF) has become a standard approach for aligning large language models with human preferences and values \cite{christiano2017deep, leike2018scalable}. RLHF emerged as a solution to the challenge of aligning AI systems with human values and preferences when these values were difficult to specify mathematically yet easy to judge. While RLHF requires relatively small amounts of comparison data to be effective compared to other approaches, sourcing high-quality preference data remains an expensive process. This technique has become particularly crucial for LLMs, where it helps guide these powerful systems toward producing outputs that humans find helpful, harmless, and honest \cite{bai2022training, bai2022constitutional}. 

The RLHF process typically consists of three stages: 
\begin{enumerate}
    \item \textbf{Supervised Fine-Tuning (SFT):} The model is first fine-tuned on demonstrations that exemplify desired behavior, producing a model $\pi^{\text{SFT}}$. 
    
    \item \textbf{Reward Modeling:}  Human annotators compare model responses, and these comparisons train a reward model $r_\phi(x, y)$ that predicts human preferences. The reward model is trained using maximum likelihood on preference pairs $(x, y_w, y_l)$ using the Bradley-Terry Model \cite{bradley1952rank, plackett1975analysis} to model the preference probability. 

    \item \textbf{RL Optimization:} The language model is then optimized further using reinforcement learning, typically with Proximal Policy Optimization (PPO), to maximize the reward while maintaining proximity to the reference model \cite{jaques2017sequence, jaques2020human, schulman2017proximal}.
\end{enumerate}
    
\subsection{Preference Optimization Methods}
Recent work has introduced more efficient alternatives to the full RLHF pipeline. Direct Preference Optimization (DPO) \cite{rafailov2023direct} eliminates the need for an explicit reward model and RL training by directly optimizing a policy from preference data:

% \begin{equation*}
% \mathcal{L}_{\text{DPO}}(\pi_\theta; \pi_{\text{ref}}) = -\mathbb{E}_{(x,y_w,y_l)\sim\mathcal{D}}\left[\log\sigma\left(\beta(r_w - r_l)\right)\right]
% \end{equation*}

\begin{equation*}
\mathcal{L}_{\text{DPO}} = -\mathbb{E}_{(x,y_w,y_l)\sim\mathcal{D}} \left[ \log\sigma\big(\beta(r_w - r_l)\big) \right]
\end{equation*}

\noindent where $r_w$ and $r_l$ are the log probability ratios of the preferred and dispreferred responses relative to a reference model. This approach has shown comparable or superior performance to RLHF while being more computationally efficient and stable.

More recent methods include Simple Preference Optimization (SimPO) \cite{meng2024simpo}, which eliminates the need for a reference model while maintaining strong performance:

\begin{equation*}
\mathcal{L}_{\text{SimPO}} = -\mathbb{E}_{(x,y_w,y_l)\sim\mathcal{D}}\left[\log\sigma\left(\beta(s_w - s_l) - \gamma\right)\right]
\end{equation*}

\noindent where $s_w$ and $s_l$ are length-normalized log probabilities, $\beta$ controls preference signal strength, and $\gamma$ is a target margin.

We also compare with Odds Ratio Preference Optimization (ORPO) \cite{Hong2024ORPOMP}, which combines supervised fine-tuning with preference optimization through a log odds ratio term, enabling effective alignment without a reference model. ORPO's loss function balances a supervised term for the preferred completion with a preference term based on log odds ratios.

\section{Data Diversification Methods}
\label{sec:data-strats}
In this section, we describe our proposed data diversification strategies on creating high-quality preference data for fine-tuning and preference optimization. 
% We evaluate the four different strategies starting from a baseline means and then utilizing existing frameworks to create data generation pipelines. 

\subsection{Baseline Strategy}
Our baseline strategy follows standard practice in preference optimization, generating multiple completions from the base model with only temperature sampling for diversity. During generation, we set the \texttt{max\_tokens} to 1,024, the \texttt{temperature} to 2, \texttt{top\_p} to 0.75, and \texttt{top\_k} to 50. We generate 5 completions from the base model $\pi^{\text{SFT}}$ using the following system prompt template:

\begin{quote}
``You will be given a math problem. Provide a step-by-step solution, clearly showing all calculations and reasoning. Ensure that each step is explained and justified. After your detailed solution, on a new line, give the final numerical answer in the format: `Final Answer: [number]'. Do not include any units in the final answer. Double-check your calculations to ensure accuracy.''
\end{quote}

\subsection{Chain-of-Thought Strategy}
Chain-of-Thought (CoT) prompting \cite{wei2022chain, kojima2022large} encourages LLMs to generate step-by-step reasoning before producing a final answer. This approach has shown significant improvements in mathematical problem-solving, particularly for complex multi-step problems \cite{havrilla2024glore}. 
For generation, we used OptiLLM's cot-reflection inference proxy to illicit chain of thought reasoning for our model during inference time\footnote{\url{https://github.com/codelion/optillm/blob/main/optillm/cot_reflection.py}}. This method implements chain-of-thought reasoning with <thinking>, <reflection>, and <output> section tags in the prompt. We set our \texttt{temperature} to 0.7 and \texttt{max\_tokens} to 1,024 to avoid context length issues with increased token counts from chain-of-thought.

\subsection{MCTS Strategy}
Methods incorporating search algorithms like Monte Carlo Tree Search (MCTS) have shown promise for enhancing mathematical reasoning \cite{feng2023alphazero, yao2023tree, liu2024mcts}. These approaches explore multiple solution paths and can identify effective reasoning strategies through simulation. For this strategy, we leverage MCTS through the OptiLLM inference proxy \cite{optillm2024} to systematically explore the solution space\footnote{\url{https://github.com/codelion/optillm/blob/main/optillm/mcts.py}}.
For each mathematical problem, we initialize a dialogue-based MCTS search with the problem as the initial query and a structured solution prompt as the system prompt. We set our \texttt{exploration\_weight} to 0.2, \texttt{num\_simulations} to 2, and our \texttt{simulation\_depth} to 1, which is the default configuration for the MCTS approach, and set \texttt{temperature} to 0.7 and \texttt{max\_tokens} to 1,024 for our generation configuration. At the end, the $N$ (in our case 5) most promising complete solution paths are picked.

This lightweight MCTS approach enables efficient yet effective exploration of the solution space, finding diverse high-quality solutions that may not be discovered through simpler sampling approaches.

\subsection{Diversified-ThinkSolve (DTS) Strategy}
\label{sec:dts}
While the previously described strategies offer certain improvements, they exhibit key limitations in generating truly diverse mathematical reasoning approaches. Temperature sampling produces variations that often follow similar reasoning patterns, and Chain-of-Thought, despite encouraging step-by-step reasoning, tends to converge on a single solution path. MCTS explores alternative branches but incurs substantial computational costs. To address these limitations, we introduce Diversified-ThinkSolve (DTS), a novel strategy specifically designed to generate diverse, high-quality mathematical reasoning paths with minimal computational overhead.

DTS leverages DSPy, a declarative programming paradigm for language models, that enables modular and structured reasoning \cite{khattab2023dspy, khattab2023dspy-assertions}\nocite{gallo2024comparative}. Unlike traditional prompting approaches that produce variations of the same solution or chain-of-thought strategies that follow a single reasoning flow, DTS explicitly decomposes the mathematical problem-solving process into two distinct phases: multiple approach generation followed by targeted execution. This decomposition enables systematic exploration of the solution space while maintaining reasoning coherence.

We implement DTS through two specialized modules. First, a \texttt{ThoughtGenerator} construct generates $N=5$ distinct reasoning approaches using the following prompt template:
\begin{quote}
``Given the math problem: \{problem\}, provide 5 possible approaches or initial thoughts on how to solve it, including any relevant mathematical concepts, formulas, or techniques that may be applied. Consider multiple perspectives and potential solution paths, and describe each thought in 1-2 sentences.''
\end{quote}

Then, for each generated approach, a \texttt{SolutionGenerator} module produces a complete solution following that particular reasoning pathway:
``Given the math problem: \{problem\} Using this approach: \{approach\} Please provide a detailed solution showing all work and steps.''

\begin{figure}[t]
    \centering
    \includegraphics[width=\columnwidth]{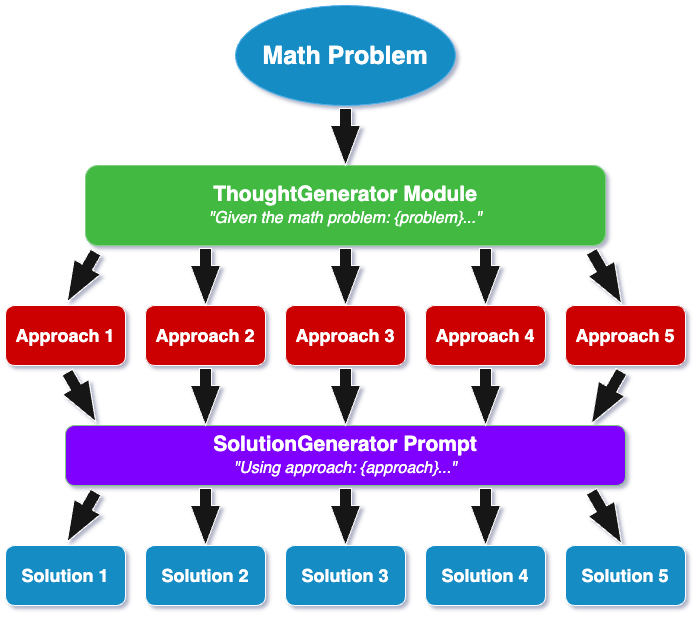}
    \caption{Diversified-ThinkSolve (DTS) modular reasoning pipeline for generating diverse mathematical problem solutions. Each math problem is first processed by a ThoughtGenerator to propose multiple solution approaches. Then utilizing the SolutionGenerator with each approach, we are given multiple solutions, contributing to a diverse set of preference data.}
    \label{fig:dspy-pipeline}
\end{figure}

Figure \ref{fig:dspy-pipeline} illustrates this process. By decoupling reasoning approach generation from solution implementation, DTS systematically explores diverse problem-solving strategies while ensuring each solution maintains consistent reasoning flow. The modularity of this approach allows for easy adjustment of the number and type of reasoning paths without modifying the entire pipeline. 

A key advantage of DTS over other strategies is its explicit promotion of strategic diversity—it doesn't merely produce different ways to present the same solution, but fundamentally different problem-solving approaches. This structured diversity creates more informative preference pairs that expose the model to a richer set of mathematical reasoning patterns during alignment.

\section{Experiment Setup}
\subsection{Datasets and Models}
We conduct our experiments using the MetaMathQA dataset \cite{yu2023metamath} composed of 395k training examples which are all augmented from the training sets of GSM8K and MATH. Since this dataset consists of duplicated problems with distinct queries, we decided to use a deduplicated version containing 13,929 unique mathematical queries and solutions. For evaluation, we use GSM8K's test set of 1,319 problems and the MATH-500 test subset. 

For training, we use \textit{meta-llama/Llama-3.1-8B-Instruct} \cite{llama31-8b-instruct} as our base model for all experiments. For scoring completions, we use Nvidia's \textit{Llama-3.1-Nemotron-70B-Reward-HF} \cite{nvidia-llama31-nemotron-70b-reward}, which demonstrated the highest accuracy in our reward model evaluation (Section \ref{sec:reward-model-selection}).

\subsection{Reward Model Selection}
\label{sec:reward-model-selection}
We evaluated several candidate reward models from the top models on RewardBench \cite{lambert2024rewardbench} by having them score both model-generated completions and ground truth solutions on the GSM8K test set. We tracked four key metrics: \texttt{correct\_higher} (model's correct output received higher reward than ground truth), \texttt{correct\_lower} (model's correct output received lower reward than ground truth), \texttt{incorrect\_higher} (model's incorrect output received higher reward than ground truth), and \texttt{incorrect\_lower} (model's incorrect output received lower reward than ground truth). An effective reward model should minimize \texttt{incorrect\_higher} cases, which represent instances where incorrect solutions are scored above correct ones.

\begin{figure}[htbp]
  \centering
  \resizebox{\linewidth}{!}{%
    \includegraphics{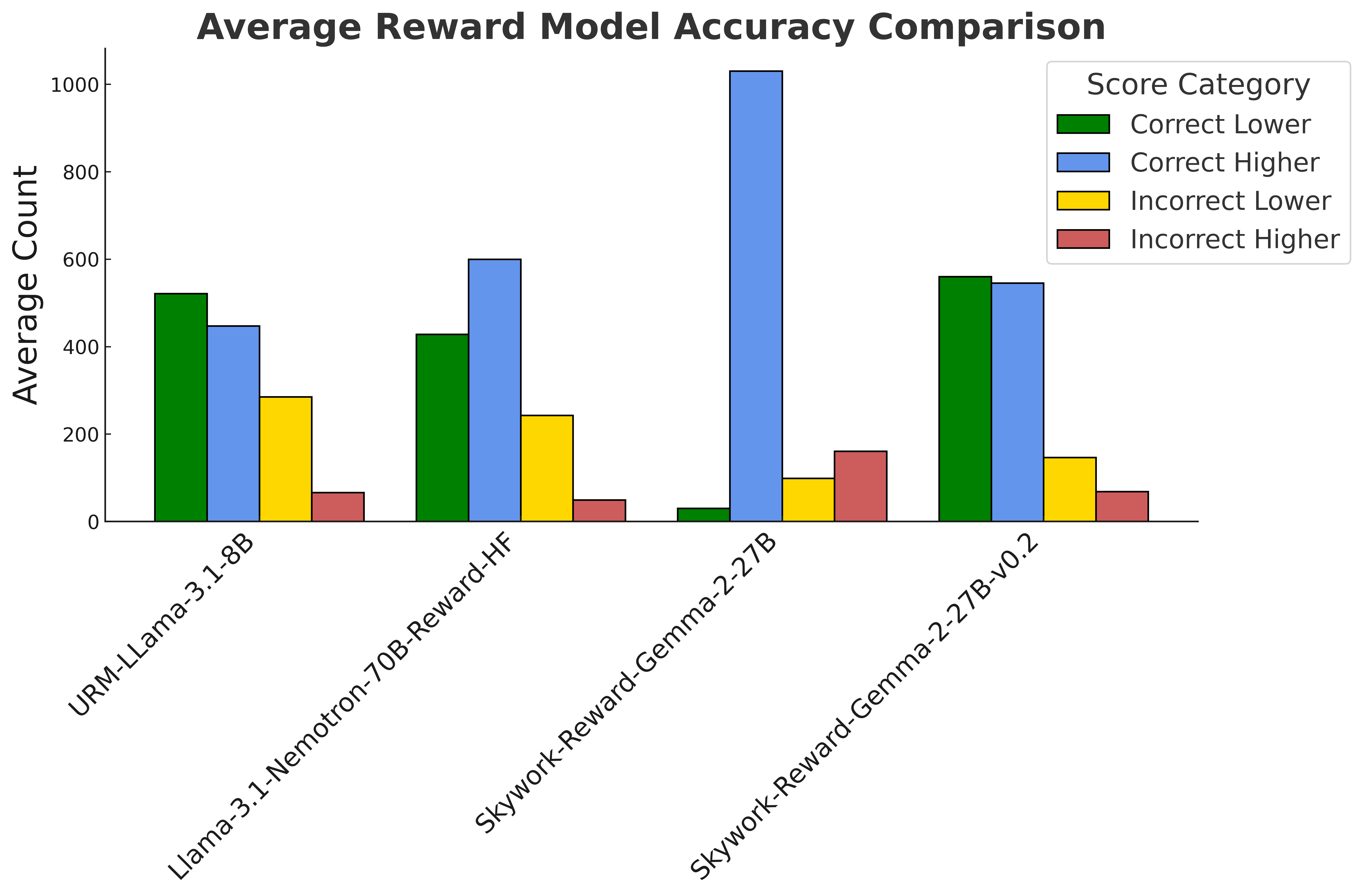}
  }
  \caption{Reward Model Accuracy Comparison. Bars represent average counts of prediction outcomes for different reward models.}
  \label{fig:reward-model-comparison1}
\end{figure}

As shown in Figure \ref{fig:reward-model-comparison1}, \textit{Nvidia's Llama-3.1-Nemotron-70B-Reward-HF} demonstrated the lowest rate of \texttt{incorrect\_higher} judgments, with an average inaccuracy rate of 3.11\% on the GSM8K test set. We further validated this model on a random sample of 4,919 problems from the harder MetaMathQA dataset, finding a comparable inaccuracy rate of 2.76\%. More details can be found in Appendix \ref{sec:reward-model-analysis}.

\subsection{Preference Data Generation and Filtering}
For each data generation strategy described in Section \ref{sec:data-strats}, we generated preference data from the MetaMathQA dataset. We applied a "mixed correctness" filtering approach, selecting only cases where 2-3 out of 5 model generations were correct, ensuring the model learns to distinguish between correct and incorrect reasoning patterns. We then used our reward model to select the highest-scored correct completion as $y_w$ and the highest-scored incorrect completion as $y_l$ for preference training.

For each strategy, we created preference datasets of comparable size: Baseline (1,097 samples, 30.4\% filtered rate from 3,610 problems), DTS (1,293 samples, 17.2\% from 7,500 problems), Chain-of-Thought (1,247 samples subsampled from 2,493, 17.9\% from the full dataset), and MCTS (1,586 samples subsampled from 3,172, 22.8\% from the full dataset). For training consistency, we used comparable dataset sizes across all strategies, sampling half of the subsets from the larger CoT and MCTS datasets.

\section{Results}
We present results comparing our different data generation strategies across various preference optimization methods.

\subsection{Analysis of Data Generation Strategies}
\label{sec:analysis-of-data-generation}
\begin{table}[ht]
  \centering
  \resizebox{\linewidth}{!}{%
  \setlength{\tabcolsep}{2.8pt}
  \begin{tabular}{lcccc}
    \toprule
    \multirow{2}{*}{\textbf{Method}} & \multicolumn{2}{c}{\textbf{GSM8K}} & \multicolumn{2}{c}{\textbf{MATH}} \\
    \cmidrule(lr){2-3} \cmidrule(lr){4-5}
    & \textbf{Best} & \textbf{Avg} & \textbf{Best} & \textbf{Avg} \\
    \midrule
    Llama-3.1-8B-IT        & \textit{76.1\%} & –    & \textit{48.2\%} & –    \\
    Llama-3.1-70B-IT       & 85.4\% & –    & 61.6\% & –    \\
    Llama-3.2-1B-IT        & 44.9\% & –    & 23.4\% & –    \\
    Llama-3.2-3B-IT       & 73.1\% & –    & 44.8\% & –    \\
    \midrule
    Baseline+SFT        & 76.7\% & 74.2\% & 48.4\% & 47.7\% \\
    Baseline+ORPO       & 77.6\% & 76.5\% & 51.8\% & 49.1\% \\
    Baseline+DPO        & 77.6\% & 76.8\% & 52.2\% & 49.6\% \\
    Baseline+SimPO      & 80.7\% & 78.9\% & 50.8\% & 48.0\% \\
    \midrule
    CoT+SFT             & 76.7\% & 73.7\% & 50.6\% & 49.0\% \\
    CoT+ORPO            & 77.4\% & 77.1\% & 51.2\% & 48.2\% \\
    CoT+DPO             & 77.6\% & 74.5\% & 50.8\% & 48.1\% \\
    CoT+SimPO           & 77.6\% & 53.0\% & 50.4\% & 48.1\% \\
    \midrule
    MCTS+SFT            & 76.7\% & 76.6\% & 48.8\% & 47.6\% \\
    MCTS+ORPO           & 79.0\% & 77.9\% & 50.8\% & 49.0\% \\
    MCTS+DPO            & 77.8\% & 77.3\% & 49.2\% & 42.8\% \\
    MCTS+SimPO          & 78.0\% & 59.7\% & 50.6\% & 24.2\% \\
    \midrule
    DTS+SFT            & 76.7\% & 75.7\% & 49.6\% & 48.8\% \\
    DTS+ORPO           & 77.2\% & 76.8\% & 50.6\% & 49.7\% \\
    DTS+DPO            & 81.2\% & 79.3\% & \textbf{52.4\%} & \textbf{50.2\%} \\
    DTS+SimPO          & \textbf{83.2\%} & \textbf{81.5\%} & 52.0\% & 49.6\% \\
    \bottomrule
  \end{tabular}
  }
  \caption{Mathematical reasoning accuracy (\%) on GSM8K (0-shot) and MATH-500 across different data generation strategies and preference optimization methods. We report both the best and average performance across 5 epochs for the optimal hyperparameter setting for each fine-tuning method and data generation strategy.}
  \label{tab:main-results}
\end{table}

As shown in Table \ref{tab:main-results}, DTS consistently outperforms other methods, achieving a 7.1\% improvement over the base Llama-3.1-8B-IT model on GSM8K and a 4.2\% improvement on MATH. The optimal fine-tuning method varies by benchmark, with SimPO yielding the best results for GSM8K and DPO performing best for MATH.

\textbf{Baseline Strategy:} The standard approach of generating completions with temperature sampling showed moderate improvements over the base model, particularly when combined with SimPO, achieving 80.7\% accuracy on GSM8K. This suggests that even simple diversity through temperature sampling can enhance performance. However, this strategy was consistently outperformed by more sophisticated diversification methods except when paired with SimPO, indicating that temperature sampling alone provides insufficient diversity for optimal mathematical reasoning. 

\textbf{Chain-of-Thought Strategy:} CoT showed mixed results with significant stability issues, particularly with SimPO where average performance dropped to 53.0\% on GSM8K despite reasonable best-case performance (77.6\%). Analysis revealed that incorrect CoT responses often contained repetitive patterns and low-quality reasoning, creating preference pairs with extremely poor rejected completions that may have hindered learning.

\textbf{MCTS Strategy:} While MCTS showed promising results with ORPO (79.0\% on GSM8K), it exhibited considerable instability with SimPO, where performance degraded substantially across epochs (average 59.7\% on GSM8K, 24.2\% on MATH). Despite MCTS's systematic exploration capabilities, its high computational cost and inconsistent performance make it less practical than DTS.

\textbf{DTS Strategy:} 
The DTS thought-based approach demonstrated substantial improvements across all fine-tuning methods, with the highest scores in both benchmarks. By explicitly generating multiple solution approaches before solving problems, this method effectively exposes the model to diverse reasoning paths. The structured exploration of different mathematical strategies appears to provide the model with a richer learning signal during preference optimization. When combined with SimPO, this approach achieved the highest average (81.5\%) and best (83.2\%) GSM8K scores, while pairing with DPO yielded the best MATH performance (52.4\%). This suggests that decomposing mathematical reasoning into distinct thought processes creates more effective preference data for alignment. 

Interestingly, the baseline strategy with SimPO outperformed both the CoT and MCTS strategies on average, highlighting that sophisticated data generation methods must be carefully integrated with the appropriate preference optimization technique. The clear winner across both benchmarks is the DTS approach, which consistently produced high-quality, diverse preference data that translated to substantial improvements in mathematical reasoning capabilities.

\subsection{Hyperparameter Sensitivity Analysis}
\label{sec:hyperparameter-analysis}

\begin{table*}[t!]
  \centering
  % Convenience macros for highlighting
  \newcommand*{\best}[1]{\textbf{#1}}
  \newcommand*{\sota}[1]{\underline{\textbf{#1}}}
  \setlength{\tabcolsep}{4pt}
  \resizebox{\textwidth}{!}{%
  \begin{tabular}{@{}lccc*{4}{cc}@{}}
    \toprule
    \textbf{Method} & \textbf{\bm{$\eta$}} & \textbf{\bm{$\beta$}} & \textbf{\bm{$\gamma$}} & \multicolumn{2}{c}{\textbf{Baseline}} & \multicolumn{2}{c}{\textbf{CoT}} & \multicolumn{2}{c}{\textbf{MCTS}} & \multicolumn{2}{c}{\textbf{DTS}} \\
    \cmidrule(lr){5-6}\cmidrule(lr){7-8}\cmidrule(lr){9-10}\cmidrule(l){11-12}
     &  &  &  & \textbf{GSM8K} & \textbf{MATH} & \textbf{GSM8K} & \textbf{MATH} & \textbf{GSM8K} & \textbf{MATH} & \textbf{GSM8K} & \textbf{MATH} \\
    \midrule
    \multirow{2}{*}{SFT} & $1\times10^{-5}$ & --- & --- & \best{76.7} & \best{48.4} & \best{76.7} & \best{50.6} & \best{76.7} & \best{48.8} & \best{76.7} & \best{49.6} \\
     & $3\times10^{-5}$ & --- & --- & 76.7 & 48.2 & 76.7 & 48.0 & 76.7 & 48.2 & 76.7 & 49.0 \\
    \midrule
    \multirow{8}{*}{DPO} & $7\times10^{-7}$ & 0.01 & --- & 75.8 & \best{52.2*} & 76.0 & 47.8 & 76.1 & 48.2 & 80.5 & 50.6 \\
     & $5\times10^{-7}$ & 0.01 & --- & 76.9 & 51.6 & \best{77.6} & 47.6 & 77.0 & \best{49.2} & \best{81.2} & 50.4 \\
     & $3\times10^{-7}$ & 0.01 & --- & 76.9 & 50.6 & 76.8 & 50.4 & 77.0 & 48.4 & 79.2 & \best{52.4*} \\
     & $1\times10^{-7}$ & 0.01 & --- & 77.6 & 48.2 & 77.5 & 50.2 & \best{77.8} & 48.4 & 77.5 & 49.2 \\
     & $3\times10^{-7}$ & 0.05 & --- & 76.7 & 51.6 & 77.3 & 50.4 & 76.9 & 49.0 & 78.6 & 51.2 \\
     & $1\times10^{-7}$ & 0.05 & --- & \best{77.6} & 51.2 & 77.0 & \best{50.8} & 77.3 & 48.8 & 77.7 & 49.0 \\
     & $3\times10^{-7}$ & 0.1 & --- & 77.2 & 48.6 & 76.7 & 48.8 & 77.3 & 48.2 & 78.2 & 51.6 \\
     & $1\times10^{-7}$ & 0.1 & --- & 77.2 & 50.8 & 77.3 & 49.2 & 77.1 & 48.6 & 77.3 & 49.0 \\
    \midrule
    \multirow{3}{*}{ORPO} & $5\times10^{-7}$ & --- & --- & 76.7 & 49.0 & 76.7 & 51.0 & \best{79.0*} & \best{50.8*} & 76.7 & \best{50.6} \\
     & $2\times10^{-7}$ & --- & --- & \best{77.6} & \best{51.8} & 77.0 & \best{51.2*} & 77.3 & 49.0 & 76.8 & 50.4 \\
     & $7\times10^{-8}$ & --- & --- & 76.7 & 49.0 & \best{77.4} & 48.6 & 77.2 & 49.6 & \best{77.2} & 48.8 \\
    \midrule
    \multirow{6}{*}{SimPO} & $1\times10^{-6}$ & 10 & 0.3 & \best{80.7*} & 49.4 & 75.7 & 48.6 & 77.9 & 49.6 & 82.8 & 48.8 \\
     & $8\times10^{-7}$ & 10 & 0.3 & 79.5 & 49.8 & 77.3 & 46.4 & 77.6 & 48.8 & 83.2 & 49.0 \\
     & $5\times10^{-7}$ & 10 & 0.3 & 78.0 & 49.6 & 77.4 & \best{50.4} & 77.9 & 49.2 & \best{83.2*} & \best{52.0} \\
     & $8\times10^{-7}$ & 10 & 0.5 & 78.5 & 50.2 & \best{77.6*} & 49.8 & 77.6 & 47.2 & 82.9 & 50.6 \\
     & $1\times10^{-6}$ & 2.5 & 0.55 & 77.7 & \best{50.8} & 76.7 & 50.2 & 77.9 & 49.6 & 82.5 & 50.6 \\
     & $8\times10^{-7}$ & 2.5 & 0.55 & 78.2 & 49.4 & 77.5 & 49.4 & \best{78.0} & \best{50.6} & 82.5 & 47.0 \\
    \bottomrule
  \end{tabular}}
\caption{Unified hyperparameter sweep across fine--tuning methods and data--generation strategies. For every hyperparameter setting we report the best--epoch accuracy (\%) on \textbf{GSM8K} and \textbf{MATH}. The highest score for each fine-tuning method's data-generation strategy is \textbf{bold}. Overall best result for each data-generation strategy is *.}
  \label{tab:unified-hyperparams}
\end{table*}

Understanding the impact of hyperparameter choices on model performance is crucial when optimizing preference learning for mathematical reasoning. While prior work has explored hyperparameter tuning for general preference learning \cite{tang2024survey}, the unique challenges of mathematical reasoning tasks may require different optimal configurations. Additionally, different data generation strategies might interact with hyperparameters in unexpected ways, potentially requiring strategy-specific tuning. We conduct this analysis to identify the most effective hyperparameter configurations for each data generation method and to provide practical guidance for researchers applying preference optimization to mathematical domains.

As shown in Table \ref{tab:unified-hyperparams}, SimPO consistently demonstrates superior performance across most data generation strategies, particularly with the DTS approach where it achieves remarkable performance (83.2\% on GSM8K). The performance advantage of SimPO is particularly pronounced with the DTS strategy, where all hyperparameter configratuions yield strong and stable results, consistently outperforming other fine-tuning methods. Notably, for both CoT and MCTS strategies, the performance margin is more modest, and in the case of MCTS, ORPO actually provides the best results for both GSM8K (79.0\%) and MATH (50.8\%). 

\textbf{SFT:} For supervised fine-tuning, we examined learning rates of 1e-5 and 3e-5, with 2e-5 being the standard default in most SFT implementations. Our results indicate minimal differences between these learning rates on GSM8K performance, with all configurations yielding identical accuracy (76.7\%). However, the lower learning rate of 1e-5 consistently produced slightly better results on the more challenging MATH benchmark across all data generation strategies, improving performance by 0.6-2.4\%. 

\textbf{ORPO:} For ORPO, we found that the standard learning rate of 8e-6 recommended in the original work was excessive for mathematical reasoning tasks, significantly degrading model performance (see Appendix \ref{sec:ORPO-learning-rate}). Our experiments with lower learning rates (5e-7, 2e-7, and 7e-8) revealed distinct optimal configurations for different data generation strategies. For MCTS, higher learning rates performed better, while the other strategies benefited from progressively lower learning rates. The best overall ORPO performance was achieved with MCTS at a learning rate of 5e-7, yielding 79.0\% on GSM8K and 50.8\% on MATH—the highest scores for any MCTS configuration across fine-tuning methods. In our ORPO ablations, we set the $\lambda$ weighing parameter by default to 1 which remained constant. 

\textbf{DPO:} Following recent findings that lower learning rates are beneficial for reasoning-intensive domains \cite{shen2024unveiling}, we conducted a thorough grid search across learning rates (1e-7, 3e-7, 5e-7, 7e-7) and beta values (0.01, 0.05, 0.1). Our results show that smaller learning rates (e.g., 5e-7) are more suitable for mathematical reasoning, with the optimal configuration varying by data generation strategy. DPO showed particularly strong performance on the MATH benchmark, achieving the highest overall MATH scores for both baseline (52.2\%) and DTS (52.4\%) strategies, representing a 4.2\% improvement over the base model. The optimal beta value was consistently 0.01 across strategies, suggesting that a mild KL constraint is preferable for mathematical reasoning tasks.

\textbf{SimPO:} Recent work has shown that when using online data with a reward model for preference data creation, increasing beta to 10 can substantially improve performance with the right learning rate \cite{meng2024simpo}. Our results strongly support this finding, as the best hyperparameters for baseline, DTS, and CoT all featured higher beta values (10) combined with carefully tuned learning rates. We also examined $\beta=2.5$ and $\gamma=0.55$, which were found promising in the original SimPO work for Llama 3 Instruct. Interestingly, while this configuration performed well, it was consistently outperformed by the higher beta configurations. The most striking result was achieved with DTS+SimPO at $\eta=5e-7$, $\beta=10$, and $\gamma=0.3$, which produced the highest overall performance on GSM8K (83.2\%) while also maintaining strong MATH performance (52.0\%).

\subsection{Training Dynamics}
\label{sec:training-dynamics}
\begin{figure*}[t]
\centering
\begin{tabular}{ccc}
\begin{tikzpicture}[scale=0.9]
\begin{axis}[
    title={GSM8K Performance},
    xlabel={Epoch},
    ylabel={Accuracy (\%)},
    xmin=0.5, xmax=5.5,
    ymin=72, ymax=84,
    xtick={1,2,3,4,5},
    ytick={72,74,76,78,80,82,84},
    ymajorgrids=true,
    grid style=dashed,
    width=0.45\textwidth,
    height=0.4\textwidth,
]
\addplot[color=blue, mark=square, line width=1.5pt]
    coordinates {(1,76.6)(2,78.7)(3,79.2)(4,80.7)(5,79.4)};
\addplot[color=green, mark=o, line width=1.5pt]
    coordinates {(1,78.2)(2,81.4)(3,81.7)(4,83.2)(5,83.0)};
\addplot[color=red, mark=triangle, line width=1.5pt]
    coordinates {(1,77.6)(2,76.0)(3,72.6)(4,73.2)(5,72.9)};
\addplot[color=purple, mark=diamond, line width=1.5pt]
    coordinates {(1,75.8)(2,77.8)(3,78.5)(4,79.0)(5,78.4)};
\end{axis}
\end{tikzpicture}
&
\begin{tikzpicture}[scale=0.9]
\begin{axis}[
    title={MATH Performance},
    xlabel={Epoch},
    ylabel={Accuracy (\%)},
    xmin=0.5, xmax=5.5,
    ymin=46, ymax=53,
    xtick={1,2,3,4,5},
    ytick={46,47,48,49,50,51,52,53},
    ymajorgrids=true,
    grid style=dashed,
    width=0.45\textwidth,
    height=0.4\textwidth,
]
\addplot[color=blue, mark=square, line width=1.5pt]
    coordinates {(1,47.4)(2,50.2)(3,47.8)(4,50.2)(5,52.2)};
\addplot[color=green, mark=o, line width=1.5pt]
    coordinates {(1,50.4)(2,52.4)(3,48.2)(4,48.4)(5,51.6)};
\addplot[color=red, mark=triangle, line width=1.5pt]
    coordinates {(1,46.8)(2,46.4)(3,49.0)(4,51.2)(5,47.6)};
\addplot[color=purple, mark=diamond, line width=1.5pt]
    coordinates {(1,49.4)(2,49.8)(3,48.0)(4,47.2)(5,50.8)};
\end{axis}
\end{tikzpicture}
&
% Legend in a standard box like the example image
\begin{tikzpicture}[baseline]
\begin{axis}[
    scale only axis,
    width=0.2\textwidth,
    height=0.3\textwidth,
    hide axis,
    xmin=0, xmax=1,
    ymin=0, ymax=1,
    legend style={
        at={(0.5,0.38)},
        anchor=center,
        draw,
        fill=white,
        nodes={scale=0.8, transform shape},
        legend cell align={left},
    },
]
\addlegendimage{blue, mark=square, line width=1.5pt}
\addlegendentry{Baseline}
\addlegendimage{green, mark=o, line width=1.5pt}
\addlegendentry{DTS}
\addlegendimage{red, mark=triangle, line width=1.5pt}
\addlegendentry{CoT}
\addlegendimage{purple, mark=diamond, line width=1.5pt}
\addlegendentry{MCTS}
\end{axis}
\end{tikzpicture}
\end{tabular}
\caption{Performance progression across training epochs for different data generation strategies using optimal hyperparameters.}
\label{fig:epoch-progression-comparison}
\end{figure*}
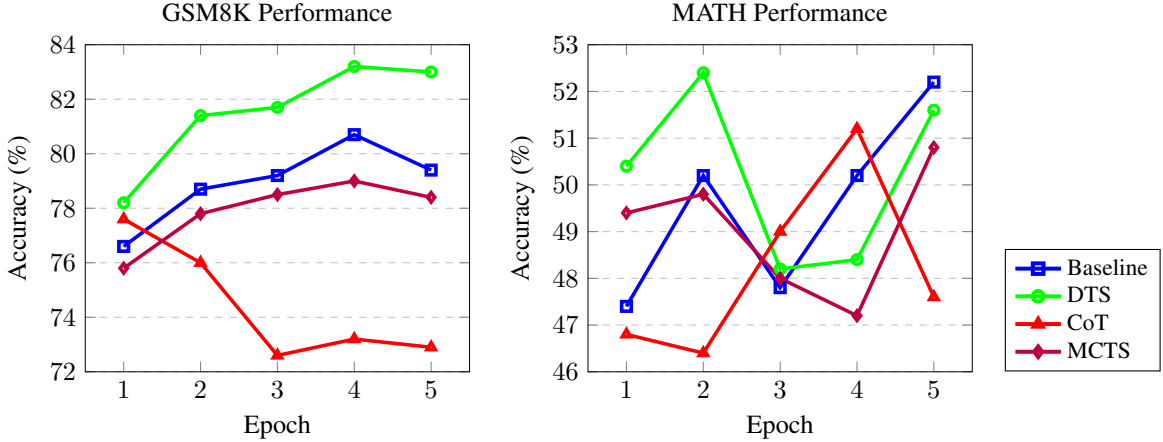

Figure \ref{fig:epoch-progression-comparison} illustrates performance evolution across training epochs for each data generation strategy. For GSM8K, we observe distinct patterns: DTS shows strong and consistent improvement, rising sharply after epoch 1 (78.2\% to 81.4\%) and maintaining growth through epoch 4 (83.2\%). In stark contrast, CoT performance degrades heavily after epoch 1, dropping from 77.6\% to 72.6\% by epoch 3, indicating significant instability. Baseline and MCTS follow similar trajectories with steady improvements until epoch 4 followed by slight regression. 

For MATH, all strategies exhibit substantial variability. Baseline and MCTS display a "dip-and-recover" pattern, with performance decreasing in the middle epochs before climbing to their peaks at epoch 5 (52.2\% and 50.8\% respectively). DTS shows similar volatility, achieving its highest performance at epoch 2 (52.4\%) before dipping and partially recovering. CoT exhibits the opposite behavior, with performance increasing until epoch 4 (51.2\%) before declining sharply at epoch 5 (47.6\%).

These dynamics highlight that DTS offers the most stable improvements for GSM8K, while all strategies demonstrate significant epoch-to-epoch variability on the more challenging MATH benchmark. 

\subsection{Computational Efficiency Considerations}
\label{sec:comp-efficiency}
Each data generation pipeline incurs a distinct token budget that corresponds to GPU hours and cost. We look at the cost of a strategy given the expected number of generated tokens per problem. The computation for the relative compute for each data generation strategy can be found in Appendix \ref{sec:token-count-estimation}. 

Despite MCTS requiring significantly more computational resources, it does not yield proportional performance improvements, failing to match either DTS or even the baseline approach with SimPO. The DTS strategy offers an exceptional balance between performance and computational efficiency with only a 1.03x compute overhead compared to baseline, making it highly suitable for resource-constrained scenarios. Even with minimal additional computation, DTS achieves the best performance on both GSM8K (83.2\%) and MATH (52.4\%).

CoT occupies a middle ground at 1.99x baseline compute, but its unstable training dynamics and inferior performance make it less attractive despite its moderate computational requirements. The baseline approach, while computationally efficient, cannot match DTS's performance despite extensive hyperparameter optimization.

\begin{table}[htbp]
  \centering
  \begin{tabular}{l>{\centering\arraybackslash}p{1.9cm}cc}
    \toprule
    \textbf{Strategy} & \textbf{Relative Compute} & \textbf{GSM8K} & \textbf{MATH} \\
    \midrule
    Baseline & 1.00x & 80.7\% & 52.2\% \\
    DTS & 1.03x & 83.2\% & 52.4\% \\
    CoT & 1.99x & 77.6\% & 51.2\% \\
    MCTS & 4.85x & 79.0\% & 50.8\% \\
    \bottomrule
  \end{tabular}
  \caption{Computational requirements and best performance for different data generation strategies combined with their respective optimal fine-tuning method.}
  \label{tab:compute-requirements}
\end{table}

\section{Conclusion}
Our findings demonstrate that strategic diversification of preference data can substantially enhance mathematical reasoning capabilities in LLMs. Several key insights emerge from our experiments:

\textbf{Diversity of reasoning paths is crucial:} Strategies that explore multiple problem-solving approaches consistently outperformed the baseline, indicating that exposure to diverse reasoning paths develops more robust mathematical capabilities.

\textbf{Data quality trumps optimization algorithm:} While SimPO and DPO performed best, the differences between optimization methods were smaller than those between data generation strategies, suggesting that research should prioritize data quality and diversity over algorithm selection.

\textbf{Structured exploration outperforms random sampling:} DTS's superior performance highlights that systematic exploration of the solution space is more effective than random variations through temperature sampling for generating high-quality preference data.

% \subsection{Future Directions}

% \qingyang{This section can be put into appendix or doesn't need to be too long}

% Several promising directions emerge from our work:

% \textbf{Enhanced DTS strategies:} Extending our approach to deeper thought hierarchies and multi-step iterations could further improve performance while maintaining computational efficiency.

% \textbf{Alternative approaches:} Exploring inference methods like OptiLLM's BON and R* reasoning, and addressing CoT's stability issues with improved filtering techniques.

% \textbf{Broader applications:} The principles underlying our approach—structured exploration of diverse reasoning paths—may transfer well to other domains that require step-by-step reasoning, including coding and logical deduction.

% Our work demonstrates that with carefully designed data generation strategies and appropriate preference optimization techniques, substantial improvements in mathematical reasoning can be achieved even with minimal computational resources.

\section{Limitations}

\subsection{Benchmark Scope and Generalizability}
Our study demonstrates improvements on GSM8K and MATH benchmarks, which, while representative, capture only a subset of mathematical reasoning tasks. The effectiveness of our strategies may vary across different mathematical domains, complexity levels, or applications. Future work should evaluate these methods on a broader range of mathematical reasoning tasks and real-world applications.

\subsection{Reward Model Dependencies}
Despite our careful selection process (achieving error rates below 3\%), our reliance on automated reward models introduces potential biases in preference data generation. These models occasionally make incorrect judgments, which could impact the quality of preference pairs and subsequent model training. Developing more robust mathematical evaluation methods remains an important avenue for future research.

\subsection{Model Scale Considerations}
Our experiments focused on a single model size (8B parameters). The relative effectiveness of different data diversification strategies might vary with model scale, potentially yielding different patterns of improvement in larger or smaller architectures. Extending this analysis to diverse model scales would provide valuable insights into the scalability of our approaches.

\subsection{Computational Efficiency Tradeoffs}
The computational requirements of more sophisticated strategies, particularly MCTS (4.85× baseline compute), limit their practical applicability in resource-constrained environments. While our DTS approach achieves an excellent balance between performance and efficiency (1.03× baseline compute), further work on optimizing data generation pipelines could improve accessibility.

\section{Ethical Considerations}

Our research aims to improve mathematical reasoning capabilities in language models, which has broadly positive applications in education, scientific research, and various technical domains.

We have made deliberate efforts to ensure research accessibility by providing comprehensive methodology details and implementation guidance. This openness helps democratize advanced mathematical capabilities across the research community and prevents the concentration of such capabilities in well-resourced organizations.

While enhanced mathematical reasoning could potentially enable more sophisticated applications in sensitive domains like finance or cryptography, we believe the educational and scientific benefits significantly outweigh potential risks. Mathematical reasoning fundamentally supports objective problem-solving rather than inherently harmful capabilities.

Our data generation methods rely on existing language models, which may contain biases. However, we focused specifically on mathematical problem-solving, which operates in a relatively objective domain with well-defined evaluation criteria, reducing (though not eliminating) the risk of perpetuating harmful biases.

We view our research as augmenting rather than replacing human mathematical reasoning, with the goal of creating more useful tools that complement human capabilities in educational and scientific contexts.

% References section
\bibliography{anthology,custom}
\bibliographystyle{acl_natbib}
\nocite{guo2024optimizing, khattab2024finetuning, zheng2023judging, chen2024assisting, gururangan2024incontext, khot2022demonstrate}

\clearpage

\appendix

\section{Additional Experimental Details}
\subsection{Implementation Details}
We implemented all models and training procedures using the HuggingFace Transformers library (version 4.43.1). For preference optimization, we used the DPO and ORPO implementations from the TRL library (version 0.9.6), which provide optimized implementations of these algorithms. All training procedures were conducted on a compute cluster with 8 NVIDIA A100 80GB GPUs using mixed-precision training (bfloat16) to accelerate training while maintaining numerical stability for mathematical operations.

For baseline model inference and data generation, we accessed the Llama-3.1-8B-Instruct model through the Together AI API \cite{togetherai2024llama31} with consistent generation parameters across experiments (unless otherwise specified). All model evaluations on the test sets were performed with greedy decoding (\texttt{temperature} = 0) to ensure deterministic outputs and fair comparisons across methods. For the different data generation strategies, we used OptiLLM (version 0.1.8) for MCTS and CoT implementations, and developed our custom DTS pipeline using core components from the DSPy framework (version 2.6.16). For reproducibility, we set random seeds consistently (42) across all experiments.

\subsection{Reward Model Analysis}
\label{sec:reward-model-analysis}

Selecting an appropriate reward model is crucial for effective preference data creation, as it directly affects the quality of paired examples used during optimization. An ideal reward model should consistently assign higher scores to correct mathematical solutions than to incorrect ones, ensuring that the preference signal aligns with mathematical accuracy.

\begin{table*}[t]
  \centering
  \setlength{\tabcolsep}{4pt}
  \begin{tabular}{llrrrrc}
    \toprule
    \multicolumn{1}{c}{\textbf{Reward Model}} & \multicolumn{1}{c}{\textbf{Generator Model}} & \multicolumn{1}{c}{\textbf{CL}} & \multicolumn{1}{c}{\textbf{CH}} & \multicolumn{1}{c}{\textbf{IL}} & \multicolumn{1}{c}{\textbf{IH}} & \multicolumn{1}{c}{\textbf{Error (\%)}} \\
    \midrule
    \multirow{4}{*}{URM-LLama-3.1-8B} & Mistral-7B-IT-v0.1 & 345 & 222 & 657 & 95 & 7.20\% \\
     & Gemma-2-9B-IT & 313 & 853 & 73 & 80 & 6.07\% \\
     & Llama-3.2-3B-IT & 691 & 349 & 235 & 44 & 3.34\% \\
     & Llama-3.1-8B-IT & 735 & 364 & 175 & 45 & 3.41\% \\
    \midrule
    \multirow{3}{*}{Llama-3.1-Nemotron-70B-Reward-HF} & Gemma-2-9B-IT & 479 & 455 & 322 & 63 & 4.78\% \\
     & Llama-3.2-3B-IT & 398 & 647 & 233 & 41 & \textbf{3.11\%} \\
     & Llama-3.1-8B-IT & 406 & 697 & 173 & 43 & 3.26\% \\
    \midrule
    \multirow{2}{*}{Skywork-Reward-Gemma-2-27B} & Llama-3.2-3B-IT & 28 & 1012 & 119 & 160 & 12.13\% \\
     & Llama-3.1-8B-IT & 32 & 1048 & 78 & 161 & 12.21\% \\
    \midrule
    Skywork-Reward-Gemma-2-27B-v0.2 & Llama-3.1-8B-IT & 560 & 545 & 146 & 68 & 5.16\% \\
    \bottomrule
  \end{tabular}
\caption{Reward Model Evaluation on the GSM8K Test Set. We evaluate various reward models against different generator models, tracking: \textbf{CL} (Correct Lower)—model's correct output received lower reward than ground truth; \textbf{CH} (Correct Higher)—model's correct output received higher reward than ground truth; \textbf{IL} (Incorrect Lower)—model's incorrect output received lower reward than ground truth; \textbf{IH} (Incorrect Higher)—model's incorrect output received higher reward than ground truth. The \textbf{Error} rate shows the percentage of incorrect outputs receiving higher rewards than ground truth, calculated as IH/(CL+CH+IL+IH).}
\label{tab:reward-model-evaluation}
\end{table*}

We conducted a comprehensive evaluation of several reward models using the GSM8K test set. For each problem, we generated solutions using various LLMs and compared the reward scores assigned to these solutions against those assigned to ground truth solutions. We tracked four key metrics, as defined in Section \ref{sec:reward-model-selection}: \texttt{correct\_lower} (CL), \texttt{correct\_higher} (CH), \texttt{incorrect\_lower} (IL), and \texttt{incorrect\_higher} (IH). The most critical metric is IH, which represents cases where an incorrect solution received a higher reward than the ground truth—these cases directly undermine the preference learning objective.

As shown in Table \ref{tab:reward-model-evaluation}, \textit{Llama-3.1-Nemotron-70B-Reward-HF} demonstrated the highest reliability, achieving the lowest error rate of 3.11\% when evaluating Llama-3.2-3B-IT outputs. The \textit{URM-LLama-3.1-8B} model also performed well with error rates below 3.5\% for the Llama-3 series, though it struggled more with Mistral-7B outputs. In contrast, the original \textit{Skywork-Reward-Gemma-2-27B} model showed the highest error rates (>12\%), frequently assigning higher rewards to incorrect solutions, though its v0.2 iteration showed substantial improvement.

To further validate our reward model selection, we extended our evaluation to the more challenging MetaMathQA dataset, sampling 4,919 problems. The Llama-3.1-Nemotron-70B-Reward-HF model maintained consistent performance with a 2.76\% error rate (136 IH cases out of 4,919 total evaluations), confirming its robustness across different mathematical problem distributions. Based on these results, we selected \textit{Llama-3.1-Nemotron-70B-Reward-HF} as our reward model for all preference data generation in our experiments.

\subsection{Judgment and Completion Scoring Setup}

Accurate assessment of mathematical solutions requires a robust scoring mechanism that can evaluate both the correctness of final answers and the quality of intermediate reasoning steps. To achieve this, we implemented a structured judgment framework using Nvidia's \textit{Llama-3.1-Nemotron-70B-Instruct-HF} model as our scoring engine.

\subsubsection{Scoring Protocol}

We normalized scores on a 0-100 scale, where 100 represents completely correct solutions with high-quality reasoning, and 0 indicates entirely incorrect solutions with flawed reasoning paths. To ensure consistent and meaningful evaluations, we designed a comprehensive system prompt that instructs the judge model to:

1. Evaluate correctness relative to reference solutions

2. Award partial credit for correct reasoning steps (up to 60 points)

3. Reserve scores of 80+ for completely correct solutions

4. Provide detailed explanations for point deductions

The full judgment prompt is structured as follows:

\begin{quote}
Here is a math question: \{question\}\\
Here is the gold answer: \{gold\_answer\}\\
Here is a student answer: \{generated\_answer\}\\
You are a math teacher grading a student's answer. You need to judge if the student answer is correct based on the gold answer. You need to follow the following rubrics:
1. The score should be between 0 and 100.
2. If the student answer is not correct based on the gold answer, deduct points from the score for each wrong step. Add points to the score for each correct step, up to a maximum of 60 points.
3. If the student answer is correct based on the gold answer, please give a final score above 80.
4. Please give a detailed explanation in bullet points for each point deducted.
In the end, the score and explanation should be in the following format. Note that the final output should be parsed as a json object.\\
<explanation>\\
\{"correct": true/false, "score": integer\}
\end{quote}

\subsubsection{Implementation Details}

To ensure scoring consistency and determinism, we set the generation parameters to \texttt{temperature=0.0} and \texttt{max\_tokens=4,096}. The structured JSON output format (\{\texttt{correct}, \texttt{score}\}) facilitated automated extraction and processing using regular expressions. In rare cases where the judge model produced malformed outputs or failed to follow the required format, we assigned a score of -1 and excluded these samples from subsequent analysis to maintain data quality.

The MetaMathQA dataset provided high-quality reference solutions that served as our gold standard for comparison. As noted in our reward model analysis (Table \ref{tab:reward-model-evaluation}), we observed occasional cases where reference solutions received lower scores than incorrect model-generated solutions. 

In our preliminary analysis, we found that this judgment approach provided more nuanced and informative scores compared to simple binary correctness checks, enabling finer distinctions between solutions with similar final answers but different reasoning quality. The detailed explanations produced by the judge model also provided valuable insights for qualitative analysis of model performance patterns and failure modes.

\subsection{Preference Optimization Configuration}

Effective preference optimization requires careful configuration of training parameters to balance learning dynamics, computational efficiency, and model stability. We implemented a consistent training infrastructure across all fine-tuning methods, varying only the specific hyperparameters detailed in our ablation study (Section \ref{sec:hyperparameter-analysis}).

\begin{figure}[t]
\begin{lstlisting}[
  basicstyle=\ttfamily\footnotesize,
  breaklines=true,
  captionpos=b,
  commentstyle=\color{green!50!black},
  frame=single,
  keywordstyle=\color{blue},
  language=bash,
  numbers=none,
  showstringspaces=false,
  stringstyle=\color{purple},
]
ACCELERATE_LOG_LEVEL=info accelerate launch \
    --config_file deepspeed_zero3.yaml \
    --dataset_name dpo_dataset \
    --model_name_or_path meta-llama/Llama-3.1-8B-Instruct \
    --learning_rate 3.0e-7 \
    --beta 0.01 \
    --lr_scheduler_type cosine \
    --bf16 true \
    --num_train_epochs 5 \
    --per_device_train_batch_size 1 \
    --gradient_accumulation_steps 16 \
    --gradient_checkpointing \
    --gradient_checkpointing_kwargs '{"use_reentrant": false}' \
    --logging_steps 25 \
    --eval_strategy 'no' \
    --optim adamw_torch \
    --attn_implementation flash_attention_2 \
    --save_strategy epoch \
    --seed 42 \
    --warmup_ratio 0.1 \
    --no_remove_unused_columns
\end{lstlisting}
\caption{Representative DPO training configuration used for fine-tuning. We maintained this base configuration across all preference optimization methods, adjusting only the method-specific hyperparameters (e.g., learning rate, beta, gamma) according to our ablation studies.}
\label{fig:dpo-command-formatted}
\end{figure}

Our training infrastructure leveraged DeepSpeed ZeRO-3 for memory optimization \cite{Rasley2020DeepSpeedSO}, FlashAttention-2 for efficient attention computation \cite{dao2023flashattention2}, and mixed-precision training (bfloat16) to accelerate training while maintaining numerical stability. We employed gradient checkpointing to reduce memory requirements, enabling us to process longer mathematical reasoning sequences without compromising batch size.

For all preference optimization methods (DPO, ORPO, SimPO), we maintained a global batch size of 16, configured as per GPU batch size of 1 with 16 gradient accumulation steps. This batch size was selected based on prior work \cite{meng2024simpo} suggesting that moderate batch sizes (16-32) achieve optimal performance for preference learning across diverse domains. Each training run was executed for 5 epochs with a cosine learning rate schedule and 10\% warmup ratio to ensure stable optimization dynamics.

Figure \ref{fig:dpo-command-formatted} shows a representative configuration for DPO training. When implementing other methods, we maintained this base configuration while adjusting method-specific parameters:

\begin{itemize}
    \item \textbf{DPO}: Varying learning rates (1e-7 to 7e-7) and beta values (0.01, 0.05, 0.1)
    \item \textbf{ORPO}: Varying learning rates (7e-8 to 5e-7) with lambda fixed at 1.0
    \item \textbf{SimPO}: Varying learning rates (5e-7 to 1e-6), beta values (2.5, 10), and gamma values (0.3, 0.5, 0.55)
\end{itemize}

For each method-strategy combination, we conducted a grid search over these hyperparameters as detailed in Section \ref{sec:hyperparameter-analysis}, totaling 76 distinct training runs. This comprehensive approach enabled us to identify optimal configurations for each data generation strategy, revealing important patterns in how hyperparameter sensitivity varies with data characteristics.

\subsection{Token Count Estimation for Computational Efficiency Analysis}
\label{sec:token-count-estimation}

To quantify the computational resources required by each data generation strategy, we developed a systematic approach for estimating relative compute costs based on token processing requirements. We use a normalized compute ratio expressed as:

\begin{equation*}
\text{Relative Compute} = \frac{t_p + t_o}{t_p^{\text{base}} + t_o^{\text{base}}}
\label{eq:rel-compute}
\end{equation*}

Where $t_p$ represents the mean prompt token count for the strategy being measured, $t_o$ is the mean output token count, and the denominator contains the corresponding values for our baseline strategy. This metric captures the computational overhead of each strategy relative to the simplest approach.

\subsubsection{Strategy-Specific Token Analysis}
\textbf{Baseline Strategy:} For our reference implementation, we measured an average problem length of 41 tokens, a system prompt of 77 tokens, and a mean generation length of 364 tokens, resulting in a total token count of $41 + 77 + 364 = 482$ tokens per problem.

\textbf{Chain-of-Thought:} Implemented using OptiLLM's Chain-of-Thought framework\footnote{\url{https://github.com/codelion/optillm/blob/main/optillm/cot_reflection.py}}, this approach uses a structured thinking template of 258 tokens combined with the problem (41 tokens), totaling 299 input tokens. We observed significantly longer generations averaging 661 tokens (due to the explicit reasoning steps and occasional verbose output), bringing the total to 960 tokens per problem. This corresponds to a compute ratio of $960/482 = 1.99\times$.

\textbf{Monte Carlo Tree Search:} Our MCTS implementation uses a simulation depth of 1 and performs 2 simulations. Based on the OptiLLM MCTS implementation\footnote{\url{https://github.com/codelion/optillm/blob/main/optillm/mcts.py}}, each problem requires a total of 8 model calls: an initial expansion, plus 4 LLM calls per simulation (generate\_actions(), apply\_action(), and evaluation\_state()) \cite{zhou2024prompt, feng2023alphazero}. With an average output of 292 tokens per model call, this strategy consumes approximately $292 \times 8 = 2,336$ tokens, yielding a compute ratio of $2,336/482 = 4.85\times$.

\textbf{Diversified-ThinkSolve (DTS):} This strategy requires two sequential LLM calls per problem:
\begin{itemize}
    \item \textit{Thought Generation:} System prompt (56 tokens) + problem (41 tokens) = 97 input tokens, producing an average of 50 output tokens
    \item \textit{Solution Generation:} System prompt (27 tokens) + problem + thought (91 tokens) = 118 input tokens, generating an average of 230 output tokens
\end{itemize}
The total token count for DTS is $(97 + 118) + (50 + 230) = 495$ tokens, resulting in a compute ratio of $495/482 = 1.03\times$ relative to baseline.

\subsubsection{Efficiency Analysis}
This token-based analysis reveals significant differences in computational requirements across strategies. While DTS achieves substantially better performance than baseline (as shown in Section \ref{sec:analysis-of-data-generation}), it does so with only a 3\% increase in computational cost. In contrast, MCTS requires nearly 5 times more compute while delivering less impressive results. These efficiency metrics provide crucial context for evaluating the practical utility of each strategy, especially in resource-constrained scenarios where computational efficiency is a key consideration alongside raw performance.

\section{Additional Results}
\label{sec:appendix-results}

\subsection{GSM8K 5-shot Performance Analysis}

While our main evaluation focused on zero-shot performance, few-shot evaluation provides valuable insights into how preference optimization affects model performance when provided with exemplars. Table \ref{tab:gsm8k-5shot} presents the GSM8K 5-shot accuracy results across all strategies and fine-tuning methods.

\begin{table}[t]
\centering
\setlength{\tabcolsep}{3.5pt}
\begin{tabular}{lcccc}
\toprule
\multicolumn{5}{l}{\textbf{Base Model (No fine-tuning)}: 83.9\%} \\
\midrule
\textbf{Strategy} & \textbf{SFT} & \textbf{DPO} & \textbf{ORPO} & \textbf{SimPO} \\
\midrule
Baseline & 84.2\% & \underline{\textbf{86.0\%}} & 85.1\% & \textbf{85.9\%} \\
CoT & 84.1\% & 85.1\% & 85.5\% & 85.3\% \\
MCTS & 84.8\% & 84.9\% & 85.3\% & 85.1\% \\
DTS & \textbf{84.8\%} & 85.8\% & \textbf{85.8\%} & 85.0\% \\
\bottomrule
\end{tabular}
\caption{GSM8K 5-shot accuracy (\%) across data generation strategies and optimization methods using optimal hyperparameter configurations. The highest score for each fine-tuning method is \textbf{bold} with the best overall result \textbf{\underline{underlined}}. All models show improvement over the base model's 83.9\% accuracy.}
\label{tab:gsm8k-5shot}
\end{table}

\subsubsection{Key Findings}

All fine-tuned models demonstrated improvements over the base model's already strong 5-shot performance (83.9\%), with gains ranging from modest (+0.2\%) to substantial (+2.1\%). Several notable patterns emerged from our analysis:

\begin{itemize}
    \item \textbf{Strategy-Method Interactions:} Unlike the 0-shot scenario where DTS consistently outperformed other strategies, the baseline strategy achieved the highest overall 5-shot accuracy (86.0\%) when combined with DPO. This suggests that the benefits of diverse reasoning paths may be partially redundant with the information provided by exemplars.
    
    \item \textbf{Method-Specific Performance:} DTS showed the most consistent performance across different fine-tuning methods, scoring strongly with SFT (84.8\%), DPO (85.8\%), and ORPO (85.8\%). However, it unexpectedly underperformed with SimPO (85.0\%) relative to other strategies, despite SimPO being the optimal method in 0-shot evaluations.
    
    \item \textbf{Hyperparameter Consistency:} We observed interesting patterns in optimal hyperparameters for 5-shot performance. For both DPO and ORPO, a learning rate of 5e-7 consistently yielded the best results across all data generation strategies, with DPO also favoring $\beta=0.01$. For SimPO, we found strategy-dependent optimal configurations: baseline, DTS, and CoT performed best with $\beta=10$, $\gamma=0.3$, and learning rates of 5e-7 or 8e-7, while MCTS uniquely benefited from $\beta=2.5$, $\gamma=0.55$, and a learning rate of 8e-7.
\end{itemize}

\subsubsection{Implications}

The differences between 0-shot and 5-shot performance patterns suggest that preference optimization may operate differently when exemplars are available. While diverse reasoning paths (as in DTS) appear critical for strong 0-shot performance, more conventional approaches like our baseline strategy may be sufficient when combined with few-shot prompting.

\subsection{Epoch-wise Analysis of MATH Benchmark Performance}

Understanding how mathematical reasoning capabilities evolve during training provides valuable insights into the learning dynamics of different preference optimization approaches. Table \ref{tab:math-epoch-progression-wide} presents a comprehensive view of MATH benchmark performance across all five training epochs for each strategy-method combination.

\begin{table*}[t]
\centering
\begin{tabular}{p{2cm}p{1.5cm}p{1.8cm}p{1.8cm}p{1.8cm}p{1.8cm}p{1.8cm}}
\toprule
\textbf{Strategy} & \textbf{Method} & \textbf{Epoch 1} & \textbf{Epoch 2} & \textbf{Epoch 3} & \textbf{Epoch 4} & \textbf{Epoch 5} \\
\midrule
\multirow{4}{*}{Baseline} 
 & SFT & \textbf{48.4\%} & 47.8\% & 47.6\% & 47.4\% & 47.2\% \\
 & DPO & 47.4\% & 50.2\% & 47.8\% & 50.2\% & \textbf{52.2\%} \\
 & ORPO & 48.8\% & 49.4\% & \textbf{51.8\%} & 47.6\% & 48.0\% \\
 & SimPO & 46.6\% & \textbf{50.8\%} & 48.4\% & 47.4\% & 47.0\% \\
\midrule
\multirow{4}{*}{CoT} 
 & SFT & 48.4\% & 48.0\% & 48.0\% & \textbf{50.6\%} & 50.0\% \\
 & DPO & 47.2\% & 47.2\% & 48.4\% & \textbf{50.8\%} & 47.0\% \\
 & ORPO & 46.8\% & 46.4\% & 49.0\% & \textbf{51.2\%} & 47.6\% \\
 & SimPO & 47.0\% & \textbf{50.4\%} & 47.6\% & 47.2\% & 48.2\% \\
\midrule
\multirow{4}{*}{MCTS} 
 & SFT & 47.8\% & \textbf{48.8\%} & 47.4\% & 47.6\% & 46.6\% \\
 & DPO & 48.2\% & \textbf{49.2\%} & 39.2\% & 39.8\% & 37.8\% \\
 & ORPO & 49.4\% & 49.8\% & 48.0\% & 47.2\% & \textbf{50.8\%} \\
 & SimPO & \textbf{50.6\%} & 42.8\% & 9.8\% & 9.0\% & 9.0\% \\
\midrule
\multirow{4}{*}{DTS} 
 & SFT & 48.0\% & 48.6\% & \textbf{49.6\%} & 48.8\% & 49.2\% \\
 & DPO & 50.4\% & \underline{\textbf{52.4\%}} & 48.2\% & 48.4\% & 51.6\% \\
 & ORPO & 48.0\% & \textbf{50.6\%} & 49.8\% & 49.4\% & 50.6\% \\
 & SimPO & 48.8\% & 51.2\% & 47.8\% & \textbf{52.0\%} & 48.0\% \\
\bottomrule
\end{tabular}
\caption{MATH benchmark accuracy (\%) progression across training epochs for all data generation strategies and optimization methods. Results shown represent the best hyperparameter configuration for each strategy-method pair. The highest score for each combination is highlighted in \textbf{bold}, while the overall best result is \underline{\textbf{underlined}}.}
\label{tab:math-epoch-progression-wide}
\end{table*}

Our epoch-by-epoch analysis reveals distinct training patterns across different approaches:

\textbf{Baseline Strategy:} While SFT performance gradually declined with additional epochs, preference optimization methods showed non-monotonic improvement patterns. Most notably, DPO achieved its peak performance (52.2\%) at the final epoch, demonstrating continued learning throughout training. Both ORPO and SimPO reached their peak performance in earlier epochs (epochs 3 and 2, respectively) before beginning to decline, suggesting potential overfitting.

\textbf{Chain-of-Thought (CoT):} Interestingly, CoT methods consistently reached their peak performance at later epochs (typically epoch 4) compared to other strategies. This delayed optimization might suggest that extracting useful signals from CoT-generated preferences requires more training time, perhaps due to the additional reasoning steps that must be learned.

\textbf{Monte Carlo Tree Search (MCTS):} MCTS exhibited the most unstable training dynamics, particularly when combined with DPO and SimPO. While MCTS+SimPO started strongly (50.6\% at epoch 1), it catastrophically collapsed to single-digit performance by epoch 3. Similarly, MCTS+DPO declined from 49.2\% to below 40\% in later epochs. This instability suggests that preferences generated through MCTS may contain conflicting or inconsistent signals that become increasingly problematic with continued training.

\textbf{Diversified-ThinkSolve (DTS):} The DTS strategy demonstrated remarkable stability across training epochs, with all methods maintaining strong performance throughout. The combination of DTS with DPO achieved the highest overall MATH accuracy (52.4\%) at epoch 2, followed by a temporary decline and subsequent recovery in later epochs. SimPO exhibited a similar pattern with its peak (52.0\%) at epoch 4. This oscillatory behavior might indicate that models trained on diverse reasoning paths explore different regions of the solution space during training.

\subsection{ORPO Learning Rate Sensitivity Analysis}
\label{sec:ORPO-learning-rate}

\begin{table}[H]
\centering
\renewcommand{\arraystretch}{1.2}
\begin{tabular}{ccc}
\toprule
\textbf{Learning Rate} & \textbf{GSM8K 0-shot} & \textbf{MATH} \\
\midrule
8e-6 & 45.6\% & 46.4\% \\
2e-6 & 68.4\% & 48.2\% \\
7e-7 & \textbf{76.9\%} & \textbf{46.8\%} \\
\bottomrule
\end{tabular}
\caption{Impact of learning rate on ORPO performance using the baseline data generation strategy.}
\label{tab:orpo-lr-sensitivity}
\end{table}

While most preference optimization methods are known to be sensitive to learning rate selection, ORPO deserves special attention due to the significantly higher learning rates recommended in the original paper (8e-6) compared to our optimal findings. We conducted targeted experiments to quantify this sensitivity and determine appropriate learning rate ranges for mathematical reasoning tasks.

As shown in Table \ref{tab:orpo-lr-sensitivity}, ORPO's performance exhibits extreme sensitivity to learning rate selection when applied to mathematical reasoning tasks. Using the originally recommended learning rate of 8e-6 results in catastrophically poor performance on GSM8K (45.6\%), significantly worse than even the untuned base model (76.1\%). Reducing the learning rate by approximately an order of magnitude (to 7e-7) restores and slightly enhances performance (76.9\%).

This stark difference can be attributed to the unique characteristics of mathematical reasoning tasks compared to general instruction-following or conversational benchmarks. Mathematical reasoning typically requires precise manipulation of symbols and strict adherence to formal rules, which may be disrupted by aggressive parameter updates. The optimal learning rate we identified (7e-7) is approximately 11 times smaller than the originally recommended value, highlighting the importance of domain-specific hyperparameter tuning.

Interestingly, while GSM8K performance was dramatically affected by learning rate selection, MATH benchmark results showed more modest variation (46.4\% to 48.2\%). This differential sensitivity might be explained by the higher complexity and diversity of problems in the MATH dataset, which may require more substantial model adaptation regardless of learning rate.

These findings emphasize the critical importance of carefully tuning learning rates when applying preference optimization techniques to specialized domains like mathematical reasoning, rather than relying on default values established for general-purpose language tasks.

\section{Analysis of Generated Solutions}
\label{sec:appendix-solutions}

To provide qualitative insights into how different data generation strategies influence mathematical reasoning patterns, we examined solution characteristics across models trained with each approach. This section presents a comparative analysis of solutions to an illustrative problem from the MetaMathQA dataset.

\subsection{Example Solutions}

\begin{quote}
\textbf{Problem:} Diego baked 12 cakes for his sister's birthday. Donald also baked 4 cakes, but ate x while waiting for the party to start. There are 15 cakes left. What is the value of unknown variable x?
\end{quote}

\textbf{Baseline Solution:}
\begin{quote}
To find the value of the unknown variable $x$, let's follow these steps:

\noindent
\textbf{Step 1:} Find the total number of cakes initially made by both Diego and Donald.

\noindent
We are given that Diego baked 12 cakes, and Donald baked 4 cakes.

\noindent
So, $\text{total\_cakes} = \text{cakes\_baked\_by\_Diego} + \text{cakes\_baked\_by\_Donald} = 12 + 4 = 16$

\noindent
\textbf{Step 2:} Understand the final number of cakes left.

\noindent
We know that after Donald ate $x$ cakes, 15 cakes are left.

\noindent
So, the initial total number of cakes minus the number eaten is equal to 15:

\noindent
$\text{total\_cakes} - x = 15$

\noindent
Since $\text{total\_cakes} = 16$ from Step 1, we substitute 16 for total\_cakes:

\noindent
$16 - x = 15$

\noindent
\textbf{Step 3:} Find the value of the unknown variable $x$.

\noindent
Now we will isolate $x$ by getting $x$ on one side of the equation:

\noindent
$16 - x = 15$

\noindent
We need to solve for $x$: $x = 16 - 15$

\noindent
\textbf{Step 4:} Perform the operation to isolate $x$.

\noindent
Using simple arithmetic: $x = 1$

\noindent
\textbf{Final Answer:} 1
\end{quote}

\textbf{Chain-of-Thought (CoT) Solution:}
\begin{quote}
\noindent
\textbf{Thinking:}

\noindent
Let's denote the total number of cakes initially as $T$.

\noindent
Donald initially had 4 cakes and ate $x$ cakes. 
Therefore, Donald had $4 - x$ cakes left.

\noindent
Diego initially had 12 cakes, so after Donald ate some, the total number of cakes left is the sum of the remaining cakes from both Diego and Donald.
The equation can be written as: $12 + (4 - x) = 15$.

\noindent
First, let's simplify the equation by combining the constants: 
$12 + 4 - x = 15$.

\noindent
Combine the constants: 
$16 - x = 15$.

\noindent
Now, let's isolate the variable $x$ by subtracting 16 from both sides of the equation.
$-x = -1$.

\noindent
To isolate $x$, multiply both sides of the equation by $-1$.
$x = 1$.

\vspace{0.5em}
\noindent
\textbf{Reflection:}

\noindent
I have simply applied basic algebraic manipulations to the given equation. There are no obvious errors in this process, and I have checked the steps for consistency and accuracy. However, it's essential to ensure that all variables are correctly isolated and that the algebraic manipulations are accurately performed.

\vspace{0.5em}
\noindent
\textbf{Output:}

\noindent
$x = 1$.
\end{quote}

\textbf{Monte Carlo Tree Search (MCTS) Solution:}
\begin{quote}
\noindent
To find the value of $x$, we need to determine the total number of cakes initially. Diego baked 12 cakes and Donald baked 4 cakes. However, since the information about how many cakes Donald initially had baked was not required for solving the problem you can just sum up Diego and Donald's initial cakes to determine the total.

\noindent
Initial total cakes = Diego's cakes + Donald's cakes = $12 + 4 = 16$ cakes.

\noindent
After waiting for the party to start, there are 15 cakes left.

\noindent
We subtract the number of remaining cakes from the total number of cakes to find $x$. This equation would be:

\noindent
Initial total cakes $-$ remaining cakes $= x$.

\noindent
$16 - 15 = x$

\noindent
$x = 1$.

\noindent
The value of the unknown variable $x$ is 1. Donald ate 1 cake while waiting for the party to start.
\end{quote}

\textbf{Diversified-ThinkSolve (DTS) Solution:}
\begin{quote}
\noindent
\textbf{Thought Output:} We can use the concept of inverse operations to isolate $x$. Subtracting 12 from both sides of the equation gives us: $-x = -5$.

\vspace{0.5em}
\noindent
\textbf{Solution Output:} To find the value of $x$, we need to isolate $x$ by using inverse operations. The correct equation is $16 - x = 15$. Subtracting 16 from both sides of the equation gives us $-x = -1$. Multiplying both sides of the equation by $-1$ gives us $x = 1$.
\end{quote}

\subsection{Qualitative Analysis of Solution Characteristics}

Our examination of solutions generated across different data strategies reveals distinctive patterns in reasoning approaches, solution structures, and error patterns.

\textbf{Baseline Solutions} typically exhibit a highly structured, step-by-step approach with explicit enumeration of each reasoning stage. The solution organization appears optimized for instructional clarity, with distinct sections and a formal problem-solving framework. While effective, this approach sometimes leads to unnecessarily verbose explanations even for straightforward problems.

\textbf{CoT Solutions} feature extensive explanatory content with distinct thinking and reflection phases. The thinking phase often includes variable definitions and elaborate equation formulations, while the reflection phase provides meta-analysis of the solution approach. This structure appears to prompt deeper verification and error-checking, but sometimes at the cost of parsimony. The explicit verification step may contribute to CoT's inconsistent performance observed in our quantitative results.

\textbf{MCTS Solutions} exhibit a remarkably consistent structure across problems, typically beginning with a standardized phrase ("To find the value of x, we need to determine...") that suggests convergence toward optimal response templates through the search process. The solutions tend to be direct and focused on the most efficient path discovered during tree search. However, this approach occasionally leads to incorrect convergence on harder problems when the search depth is insufficient to fully explore the solution space.

\textbf{DTS Solutions} demonstrate a unique two-phase structure reflecting the strategy's decomposition approach. The initial "thought output" often contains a high-level strategy or alternative solution approach, while the subsequent "solution output" provides a direct, efficient solution path. This dual structure appears to enable a balance between conciseness and reasoning depth. The example solution illustrates how DTS can derive a more direct mathematical approach (using inverse operations) compared to other methods.

\section{Diversified-ThinkSolve (DTS) Implementation Details}

DTS was implemented using DSPy framework components, with a modular design that separates thought generation from solution execution. Our implementation consists of two primary modules which can be found in Figure \ref{fig:thought-generator-code} and Figure \ref{fig:solution-generator-code}. The DTS implementation incorporates several key design elements:

\textbf{Modularity:} By separating thought generation from solution execution, each component can be independently optimized.

\textbf{Robust Error Handling:} Comprehensive fallback mechanisms prevent pipeline failures during batch processing.

\textbf{Structured Output Processing:} Regex-based parsing extracts individual thoughts from varied model outputs, ensuring consistent downstream processing.

\textbf{Guaranteed Diversity:} The system enforces a minimum of five distinct approaches per problem, even when the base model tends toward homogeneity.

The complete pipeline processes each problem through ThoughtGenerator, passes each generated approach to SolutionGenerator, collects the resulting solutions, scores them with the reward model, and creates preference pairs for optimization. This architecture maintains computational efficiency (1.03× baseline) while producing the diverse, high-quality preference data that enabled DTS's superior performance.

\begin{figure*}[t!]
\begin{lstlisting}[
  basicstyle=\ttfamily\footnotesize,
  breaklines=true,
  captionpos=b,
  commentstyle=\color{green!50!black},
  frame=single,
  keywordstyle=\color{blue},
  language=Python,
  numbers=left,
  showstringspaces=false,
  stringstyle=\color{purple},
]
class ThoughtGenerator(dspy.Module):
    def __init__(self):
        super().__init__()
        self.gen_thoughts = dspy.ChainOfThought("math_problem -> thoughts: List[str]")
    
    def forward(self, math_problem: str) -> List[str]:
        try: 
            prompt_template = (
                "Given the math problem: {problem}, provide 5 possible approaches or "
                "initial thoughts on how to solve it, including any relevant mathematical "
                "concepts, formulas, or techniques that may be applied. Consider multiple "
                "perspectives and potential solution paths, and describe each thought in 1-2 sentences."
            )
            
            result = self.gen_thoughts(math_problem=prompt_template.format(problem=math_problem))
            thoughts = result.reasoning if hasattr(result, 'reasoning') else []
                
            # process thoughts
            if isinstance(thoughts, str):
                import re
                thoughts = re.split(r'\d+\.|\n\d+\.|\n\d+\)', thoughts)
                thoughts = [t.strip() for t in thoughts if t.strip()]
            
            # ensure exactly 5 thoughts
            if len(thoughts) < 5:
                while len(thoughts) < 5:
                    thoughts.append(f"Alternative approach {len(thoughts) + 1}: Apply fundamental mathematical principles to solve step by step.")
            
            return thoughts
        
        except Exception as e:
            print(f"Error in ThoughtGenerator: {str(e)}")
            return [f"Default approach {i+1}: Solve the problem systematically using basic mathematical principles." 
                   for i in range(5)]
\end{lstlisting}
\caption{ThoughtGenerator module implementation responsible for generating diverse mathematical reasoning approaches.}
\label{fig:thought-generator-code}
\end{figure*}

\begin{figure*}[t!]
\begin{lstlisting}[
  basicstyle=\ttfamily\footnotesize,
  breaklines=true,
  captionpos=b,
  commentstyle=\color{green!50!black},
  frame=single,
  keywordstyle=\color{blue},
  language=Python,
  numbers=left,
  showstringspaces=false,
  stringstyle=\color{purple},
]
class SolutionGenerator(dspy.Module):
    def __init__(self):
        super().__init__()
        self.gen_solution = dspy.ChainOfThought("math_problem, approach -> solution: str")
    
    def forward(self, math_problem: str, approach: str) -> str:
        try:
            prompt_template = (
                "Given the math problem: {problem}\n"
                "Using this approach: {approach}\n"
                "Please provide a detailed solution showing all work and steps."
            )
            
            result = self.gen_solution(math_problem=math_problem, approach=approach)

            return result.reasoning if hasattr(result, 'reasoning') else "Unable to generate solution."
            
        except Exception as e:
            print(f"Error in SolutionGenerator: {str(e)}")
            return "Error occurred while generating solution."
\end{lstlisting}
\caption{SolutionGenerator module implementation that produces complete solutions based on specific reasoning approaches.}
\label{fig:solution-generator-code}
\end{figure*}

\end{document}